\documentclass{IOS-Book-Article}

\usepackage{mathptmx}
\usepackage{soul}\setuldepth{article}
\usepackage{comment}
\usepackage{graphicx,array}
\usepackage[T1,T5]{fontenc} 
\usepackage{multirow}
 
\def\hb{\hbox to 10.7 cm{}}

\begin{document}

\pagestyle{headings}
\def\thepage{}

\begin{frontmatter}              

\title{Facial Expression Recognition and Image Description Generation in Vietnamese}


\author[A]{\fnms{Khang Nhut} \snm{LAM}
\thanks{Corresponding Author: Khang Nhut Lam, Can Tho University, Campus II, 3/2 Street, Can Tho City, Vietnam; E-mail: lnkhang@ctu.edu.vn.}},
\author[A]{\fnms{Kim-Ngoc Thi} \snm{NGUYEN}},
\author[A]{\fnms{Loc Huu} \snm{NGUY}},

and
\author[B]{\fnms{Jugal} \snm{KALITA}}

\address[A]{Can Tho University, Can Tho, Vietnam}
\address[B]{University of Colorado, Colorado Springs, USA}

\begin{abstract}
 This paper discusses a facial expression recognition model and a description generation model to build descriptive sentences for images and facial expressions of people in images. Our study shows that YOLOv5 achieves better results than a traditional CNN for all emotions on the KDEF dataset. In particular, the accuracies of the CNN and YOLOv5 models for emotion recognition are 0.853 and 0.938, respectively. A model for generating descriptions for images based on a merged architecture is proposed using VGG16 with the descriptions encoded over an LSTM model. YOLOv5 is also used to recognize dominant colors of objects in the images and correct the color words in the descriptions generated if it is necessary. If the description contains words referring to a person, we recognize the emotion of the person in the image. Finally, we combine the results of all models to create sentences that describe the visual content and the human emotions in the images. Experimental results on the Flickr8k dataset in Vietnamese achieve BLEU-1, BLEU-2, BLEU-3, BLEU-4 scores of 0.628; 0.425; 0.280; and 0.174, respectively.
\end{abstract}

\end{frontmatter}
\markboth{K.N. Lam et al. / Facial Expression Recognition and Image Description Generation in Vietnamese\hb}{K.N. Lam et al. / Facial Expression Recognition and Image Description Generation in Vietnamese\hb}
\vspace{-0.75cm}
\section{Introduction}
\vspace{-0.25cm}
Facial expression recognition (FER) and image description generation (IDG) are important tasks in image understanding, connecting computer vision with natural language processing. Image content can be partially described through objects and their locations. Huang et al. \cite{ref_Huang2019} classify approaches for FER into 2 groups, including conventional FER and deep learning based approaches. Given an input image, the conventional FER approach consists of several steps: pre-processing the image to reduce noise (e.g., Gaussian Filter \cite{ref_Deng1993}, Bilateral Filter \cite{ref_Zhang2009}), detecting face and facial components, extracting features (e.g., Local Directional Pattern \cite{ref_Jabid2010}, Histogram of Oriented Gradients \cite{ref_Dalal2005}), and classifying emotions (e.g., Support Vector Machine \cite{ref_Eng2019}, Na\"ive Bayes classifier \cite{ref_Mao2016}). Li and Deng  \cite{ref_Li2020} report that deep FER consists of several steps including face alignment detectors (e.g., Viola-Jones \cite{ref_Viola2001}, face alignment 3000fps \cite{ref_Ren2014}), data augmentation (e.g., rotation, skew, scaling), face normalization, feature learning, and emotion classification. Several deep neural network models have been used to learn image features such as Convolutional Neural Network (CNN) \cite{ref_Walecki2017,ref_Pranav2020}, the hybrid Convolution - RNN \cite{ref_Jain2018}, and Generative Adversarial Network \cite{ref_Yang2021}. 
Human facial expressions are usually classified into 7 categories \cite{ref_Lundqvist1998}: afraid, angry, disgusted, happy, neutral, sad, and surprised. 

Publishing efforts on IDG may be grouped into 3 categories \cite{ref_Bernardi2016,ref_Hossain2019}: (i) Models rely on computer vision techniques to identify objects in the input image and extract their features \cite{ref_Kulkarni2013,ref_Elliott2013}. Features extracted are fed to a Natural Language Generation (NLG) \cite{ref_Reiter1997} sub-system. Then, the steps to build a description for the image follow the NLG architecture. (ii) Models are based on a retrieval system, where the image descriptor is retrieved from the training dataset. Most of these systems use neural models to extract image features and linguistic information \cite{ref_Socher2014,ref_Sur2020}. (iii) A system relies on the generation architecture to generate new descriptions. First, neural models are used to extract features of images (e.g., VGGNet \cite{ref_Simonyan2014}, Faster R-CNN \cite{ref_Ren2015}, ResNet \cite{ref_He2016}, and Inception-V3 \cite{ref_Szegedy2016}), then neural models 
are used to generate new descriptions \cite{ref_Aneja2018,ref_Guo2020}. 
For the last category, there are 2 architectures to generate images captions: inject and merge architectures \cite{ref_Tanti2017}. 
In the inject architecture, the vectors of image features and words are combined and fed into a neural model for generating image captions; whereas in the merge architecture, the image feature vectors are merged with the final state of the neural model in a multimodal layer. The experiments show that the merge architecture outperforms the inject architecture. 

The more information provided in the descriptive sentences, the image caption model is better. For example, the image caption ``A boy with a happy face in a red shirt is playing on grass'' is more detailed and vivid than the image caption ```A boy is playing on the grass''. We have not seen image captioning with emotion recognition. This paper aims to explore the methods for FER and IDG. If the sentence describing the image contains words or phrases referring to a human, the system will identify the facial expression and add this emotion to the image description. In other words, the description sentence describes the content in the image and the facial emotion of the person.

\vspace{-0.68cm}
\section{Proposed Approach} \vspace{-0.28cm}
We first discuss the datasets used and methods to pre-process datasets. Then, we present approaches to recognize human facial expressions and generate descriptions of images.\vspace{-0.28cm}
\subsection{Datasets Pre-processing}\vspace{-0.28cm}
The KDEF \cite{ref_Lundqvist1998}, Flickr8k \cite{ref_Hodosh2013}, and Flickr30k\cite{ref_Young2014} datasets are used to train the models to recognize the human facial expressions and to generate the image descriptions, respectively. The KDEF dataset comprises 4,900 images of 70 individuals displaying 7 emotional expressions, each of which is viewed from 5 different angles. The dataset is divided by 70\%, 20\%, and 10\% for training, validation, and test sets, respectively. The Flickr8k and Flickr30k datasets consist of 8,092 images and 31,783 images, respectively, each of which has 5 description sentences. Each dataset is divided into 2 parts: 1,000 images for testing, the rest of the images for training. 

The description sentences in the Flickr8k dataset are in English. We translate this dataset to Vietnamese using a pre-trained Transformer model\footnote{https://github.com/pbcquoc/transformer}. The Transformer translation model was trained on the dataset using 600,000
sentences extracted from TED\footnote{https://www.ted.com/}. Then, we pre-process the description sentences by converting sentences to lowercases, removing special characters and punctuation marks. The Underthesea\footnote{https://pypi.org/project/underthesea/} toolkit is used to segment words in the description sentences. The dictionary created consists of 4,028 words. Image descriptions are embedded in terms of vectors based on the position of words in the dictionary. \vspace{-0.28cm}
\subsection{CNN-based and YOLOv5-based FER Models} \vspace{-0.28cm}
The CNN model \cite{ref_LeCun1998} has 3 types of layers, including convolution, pooling, and fully connected layers. The convolution layer extracts feature maps from input images by using filters to perform convolution operations. Then, the features extracted are applied nonlinear transfer functions such as ReLU, sigmoid, tanh, and softmax. The pooling layer reduces the dimensionality of the output of the previous layer by performing a pooling operation such as max pooling, average pooling, and sum pooling. Finally, the fully connected layer or dense layer is a normal flat feed-forward neural network layer using a nonlinear activation function to obtain the probability of each class. In this paper, we use a classic feed-forward CNN to detect facial expressions. The OpenCV library is used to detect and crop human faces in each image, and then convert face images to grayscale. The grayscale images are fed to the CNN model using ReLU activation function, average pooling operation for training FER.

A traditional CNN does not detect and label objects well in real-time. YOLO \cite{ref_Redmon2016} is state-of-the-art in real-time detecting objects. 
YOLOv3 can predict the bounding box and process the image simultaneously, so it is less time-consuming. The accuracies of using YOLOv3 for facial expression recognition on JAFFE, RaFD, and CK+ are 98.12\%, 97.01\%, and 99.72\% \cite{ref_Luh2019}, respectively. Currently, the newest version YOLOv5 comprises 3 main components: Cross Stage Partial Network \cite{ref_Wang2020} (CSP) backbone (including CSPRestNext50 and CSPDarknet) for feature extraction, PA-NET \cite{ref_Wang2019} neck for feature aggregation, and Head with YOLO layer for predicting boxes and labels. In our FER experiment, we use YOLOv5 provided by Ultralytics\footnote{https://ultralytics.com/yolov5}. \vspace{-0.28cm}

\subsection{Image Description Generation Model}
\begin{figure} [h]
	\centering
	\includegraphics[width=\textwidth]	{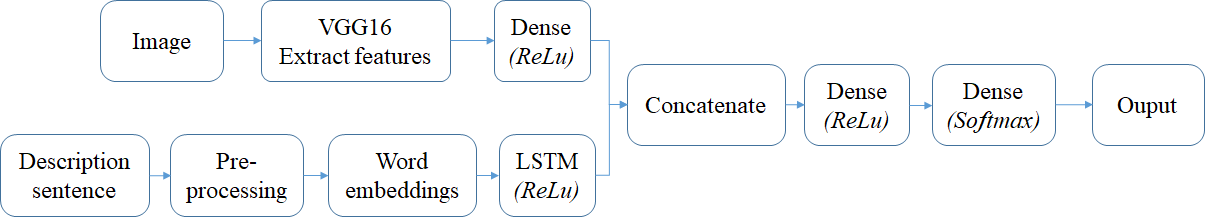} 
	\caption{The image description generation model  \cite{ref_Tanti2017,ref_Sharma2019}}  
	\label{fig:IDC}
\end{figure}

\vspace{-0.28cm}
We use the merge architecture to construct descriptions for images  \cite{ref_Tanti2017,ref_Sharma2019} with the VGG16 model for extracting image features and the LSTM model for constructing image caption, as presented in Figure~\ref{fig:IDC}. In our implementation, we use the library toolkits supported by TensorFlow. \vspace{-0.2cm}
\vspace{-0.28cm}

The VGG16 model extracts image feature vectors of size 4,096, which then are fed into a dense layer using the ReLU activation function. Description sentences are pre-processed, segmented, padded, and embedded into vectors. The word embeddings are fed to the LSTM using ReLU function. The outputs of the dense neural network and the LSTM have a similar size of 256 and then are concatenated and passed through a Dense layer using ReLU function with an output space of 256 and the next Dense layer using the softmax activation function with an output space of the dictionary size of 4,028.

If there is a personal noun in the caption generated, we simply pass the human image into the FER model and add the result of the model to the description generated. A list of person nouns is constructed manually including \{{\fontencoding{T5}\selectfont \dj\'\uhorn a tr\h{e} (kid), ng\uhorn \`\ohorn i \dj\`an \ocircumflex ng (man, male), c\d\acircumflex u b\'e (boy), thanh ni\ecircumflex n (adolescent), tr\h{e} em (baby), c\ocircumflex ~b\'e (girl), ng\uhorn\`\ohorn i ph\d{u} n\~\uhorn ~(woman, female, lady), ch\`ang trai (boy), \ocircumflex ng gi\`a (old man), b\`a gi\`a (old woman), em b\'e (baby), b\'e g\'ai (girl)}\}.

\begin{figure} [h]
	\centering
	\includegraphics[scale=0.35]		{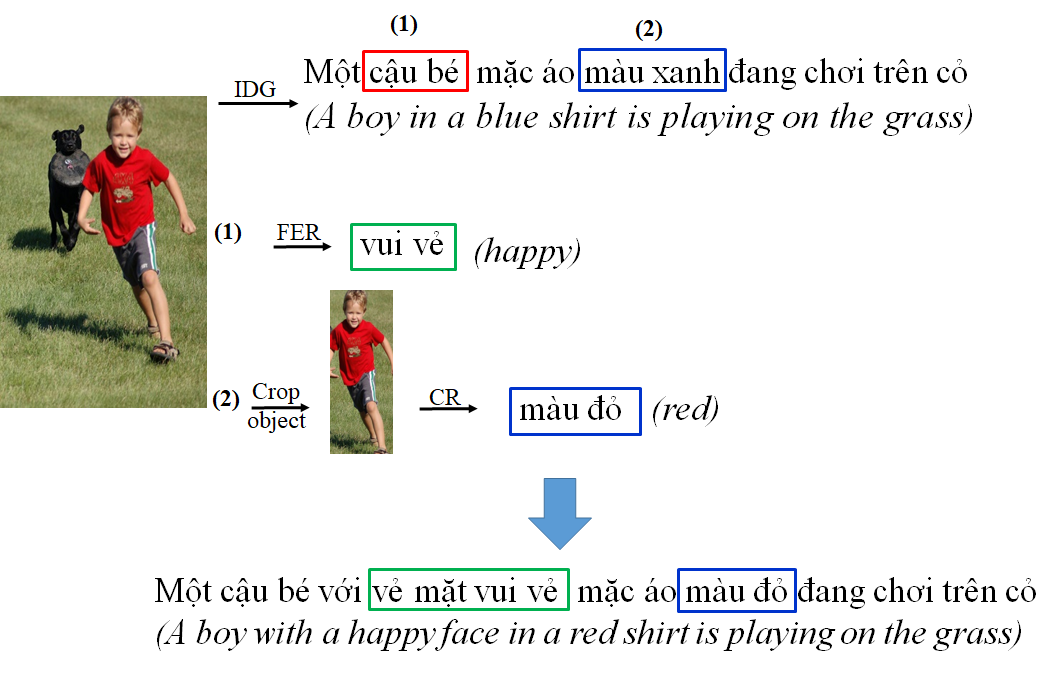} \vspace{-0.25cm}
	\caption{Example of generating an image description } \label{fig:exIDC}
\end{figure}

During the experiment, we noticed that sometimes the model did not perform correctly in determining the color of the object in the image. To solve the problem of color recognition (CR), from the built-in descriptive sentence, if the sentence contains color words, we determine the name of the object that needs color recognition. We use the YOLOv5 model to locate the object, crop the object, and save it as a new image for color recognition. An object usually has more than one color, we extract the dominant color of the object following the instructions of Ercolanelli\footnote{https://github.com/algolia/color-extractor}. When cropping an object out of an image, the object is usually in the center of the image, so the colors in the four corners of the image are usually the background colors. Therefore, pixels close to the color of the four corners of the image are considered the background color and excluded from the image. Then, a clustering algorithm, the K-means algorithm, is used to group similar pixels. The most dominant color of the object is considered the color of that object. Finally, the K Nearest Neighbors algorithm is used to convert the object color to words in human language by finding the nearest neighbor color in a large dictionary of colors taken from the XKCD color survey. The color after identification is replaced with the color of the previous description. Figure~\ref{fig:exIDC} shows an example of generating an image description.
\vspace{-0.35cm}
\vspace{-0.9cm}
\section{Experimental Results}
\vspace{-0.28cm}
We build the modes in the Google Colab environment with 12GB RAM with GPU. The FER test set has 490 images with 78 afraid (meaning ``{\fontencoding{T5}\selectfont s\d\ohorn~h\~ai}''), 83 angry (meaning ``{\fontencoding{T5}\selectfont gi\d\acircumflex n d\~\uhorn }''), 71 disgusted (meaning ``{\fontencoding{T5}\selectfont gh\ecircumflex ~t\h\ohorn m}''), 65 happy (meaning ``{\fontencoding{T5}\selectfont vui v\h{e} }''), 66 neutral (meaning ``{\fontencoding{T5}\selectfont trung l\d\acircumflex p}''), 64 sad (meaning ``{\fontencoding{T5}\selectfont bu\`\ocircumflex n b\~a}''), and 63 surprised (meaning ``{\fontencoding{T5}\selectfont ng\d{a}c nhi\ecircumflex n}'') emotions. 
The results of the FER models are presented in Table~\ref{tab:evaluateFacialEmotion}. 

\begin{table}[h]
	\centering
	\caption{Recall, precision and F1-score of the FER models}\label{tab:evaluateFacialEmotion}
	\begin{tabular}{|l|c|c|c|c|c|c|}
		\hline
		Facial &\multicolumn{2}{c}{Recall}&\multicolumn{2}{|c|}{Precision}&\multicolumn{2}{c|}{F1-score} \\ \cline{2-7}
		emotions& ~CNN~ &~ YOLOv5~ & ~CNN ~& ~YOLOv5~ &~ CNN~ &~ YOLOv5~ \\
		\hline
		Afraid & 0.705&0.897&0.785&0.921&0.742&0.908 \\
		Angry&0.795&0.939&0.942&0.975&0.862&0.956 \\
		Disgusted&0.859&0.901&0.813&0.941&0.835&0.920 \\
		Happy&0.953&0.969&0.968&0.984&0.960&0.976\\
		Neutral&0.954&0.984&0.851&0.984&0.899&0.984\\
		Sad&0.812&0.937&0.776&0.845&0.793&0.888\\
		Surprised&0.920&0.952&0.828&0.923&0.871&0.937 \\
		
		\hline
	\end{tabular}
\end{table}
 The accuracies of the CNN and YOLOv5 models are 0.853 and 0.938, respectively. YOLOv5 recognizes human facial emotions better and faster than CNN in all emotions. The happy emotion can be recognized very well; whereas, the afraid emotion might be mis-recognized as the surprise emotion by both two emotion recognition models. Next, we evaluate the method for generating image descriptions in Vietnamese. Table~\ref{tab:scoreIDG} presents the BLEU-scores of the image description generation model (the so-called IDG) and the image description generation model with facial expression recognition (the so-called IDG with FER). Interestingly, the results show that the Flickr30k dataset, including more images and captions than the Flickr8k dataset, does not help achieve better BLEU scores. Some examples of the image descriptions are shown in Table~\ref{tab: example}.\vspace{-0.28cm}
\begin{table}[h]
	\centering
	\caption{BLEU-scores of the image description generation models}\label{tab:scoreIDG}
	\begin{tabular}{|c|c|c|c|c|c|}
		\hline
		Dataset& Model &BLEU-1&BLEU-2&BLEU-3&BLEU-4 \\  \hline
		\multirow{2}{*}{Flickr8k}&IDG&0.629&0.426&0.281&0.175\\
		 &IDG with FER&0.628&0.425&0.280&0.174\\ \hline
 		\multirow{2}{*}{Flickr30k}&IDG&0.616&0.396&0.242&0.136\\
		 &IDG with FER&0.615&0.396&0.241&0.135\\
		\hline
	\end{tabular}
\end{table}\vspace{-0.28cm}

\begin{table}[h]
	\caption{Examples of the image descriptions} \label{tab: example}
	\centering
	\begin{tabular}{|m{0.25\textwidth}|m{0.25\textwidth}|m{0.35\textwidth}|}	\hline
		
		Image &  Image description & Image description with correction\\  \hline
		
		\begin{center}\vspace{-0.15cm}\includegraphics[scale =0.6]{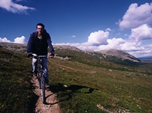}\vspace{-0.2cm}\end{center} &{\fontencoding{T5}\selectfont M\d\ocircumflex t ng\uhorn\`\ohorn i \dj\`an \ocircumflex ng \dj ang \dj i xe \dj\d{a}p tr\ecircumflex n m\d\ocircumflex t con  \dj \uhorn\`\ohorn ng} (meaning ``A man is riding a bicycle on a road'').& \begin{center} None \end{center} \\ \hline
		
		\begin{center}\vspace{-0.15cm}\includegraphics[scale =0.55]{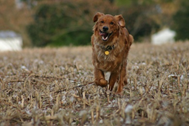}\vspace{-0.25cm}\end{center} &
		{\fontencoding{T5}\selectfont M\d\ocircumflex t con ch\'o \dj en v\`a tr\'\abreve ng \dj ang ch\d{a}y qua m\d\ocircumflex t c\'anh \dj \`\ocircumflex ng} (meaning ``A black and white dog is running through a field'').&  {\fontencoding{T5}\selectfont M\d\ocircumflex t con ch\'o n\acircumflex u \dj ang ch\d{a}y qua m\d\ocircumflex t c\'anh \dj \`\ocircumflex ng} (meaning ``A brown dog is running through a field''). \\ \hline
		
		\begin{center}\vspace{-0.15cm}\includegraphics[scale =0.55]{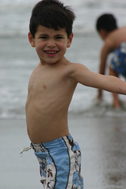}\vspace{-0.25cm}\end{center} & {\fontencoding{T5}\selectfont M\d\ocircumflex t c\d\acircumflex u b\'e \dj ang ch\ohorn i trong h\`\ocircumflex ~b\ohorn i} (meaning ``A boy is playing in the pool''). & {\fontencoding{T5}\selectfont M\d\ocircumflex t c\d\acircumflex u b\'e v\'\ohorn i v\h{e} m\d\abreve t vui v\h{e} \dj ang ch\ohorn i trong h\`\ocircumflex ~b\ohorn i} (meaning ``A boy with a happy facial expression is playing in the pool'').\\ \hline

	\end{tabular}
\end{table}\vspace{-0.58cm}
\section{Conclusion}\vspace{-0.28cm}

We experiment with CNN and YOLOv5 to recognize facial expressions on the KDEF dataset. The image description generation model integrated with the emotion recognition model and color recognition model achieves acceptable BLEU scores on the Flickr8k dataset. The description sentences generated by the current model can describe one person in an image. For future work, we will study approaches that might describe many people and their emotions in the image. Currently, we are performing experiments using Inception-V3 and YOLOv5 to extract image features instead of the VGG16 model, and training the models on different datasets. Besides, the Transformer model \cite{ref_transformers2020} or BERT model \cite{ref_bert2018} will be used to generate description sentences. In addition, we need to improve the quality of the training dataset by improving the translation model.

\end{document}